\def\year{2022}\relax
\documentclass[letterpaper]{article} 
\usepackage{aaai22}  
\usepackage{times}  
\usepackage{helvet}  
\usepackage{courier}  
\usepackage[hyphens]{url}  
\usepackage{graphicx} 
\usepackage{natbib}  
\usepackage{caption} 
\DeclareCaptionStyle{ruled}{labelfont=normalfont,labelsep=colon,strut=off} 
\usepackage{algorithm}
\usepackage{algorithmic}

\usepackage{newfloat}
\usepackage{listings}
\floatstyle{ruled}
\newfloat{listing}{tb}{lst}{}
\floatname{listing}{Listing}

\makeatletter
\newcommand{\Ni}{({\em i})~}
\newcommand{\Nii}{({\em ii})~}
\newcommand{\Niii}{({\em iii})~}

\usepackage{amssymb}
\usepackage{amsmath}
\usepackage{pifont}
\newcommand{\cmark}{\ding{51}}%
\newcommand{\xmark}{\ding{55}}%
\def\eg{\emph{e.g}.}
\def\ie{\emph{i.e}.}
\usepackage{xcolor}
\newcommand{\prj}[1]{\textcolor{black}{#1}} 
\newcommand{\jh}[1]{\textcolor{black}{#1}} 

\newcommand{\jx}[1]{\textcolor{black}{#1}} 
\usepackage{multirow}
\usepackage{array}
\makeatother

\title{UNISON: Unpaired Cross-lingual Image Captioning}
\author {
    Jiahui Gao\textsuperscript{\rm 1},
    Yi Zhou\textsuperscript{\rm 2},
    Philip L.H. Yu\textsuperscript{\rm 3}\thanks{Corresponding author.},
    Shafiq Joty\textsuperscript{\rm 4},
    Jiuxiang Gu\textsuperscript{\rm 5}
}
\affiliations {
    \textsuperscript{\rm 1} The University of Hong Kong, Hong Kong \ \textsuperscript{\rm 2} Johns Hopkins University, USA \\
    \textsuperscript{\rm 3} The Education University of Hong Kong, Hong Kong \ \textsuperscript{\rm 4} Nanyang Technological University, Singapore \\
    \textsuperscript{\rm 5} Adobe Research, USA \\
    sumiler@hku.hk, \ yzhou188@jhu.edu, \ plhyu@eduhk.hk, \ srjoty@ntu.edu.sg, \ jigu@adobe.com
}

\begin{document}

\maketitle

\begin{abstract}
Image captioning has emerged as an interesting research field in recent years due to its broad application scenarios. The traditional paradigm of image captioning relies on paired image-caption datasets to train the model in a supervised manner. However, creating such paired datasets for every target language is prohibitively expensive, which hinders the extensibility of captioning technology and deprives a large part of the world population of its benefit. In this work, we present a novel unpaired cross-lingual method to generate image captions without relying on any caption corpus in the source or the target language. Specifically, our method consists of two phases: \Ni a cross-lingual auto-encoding process, which utilizing a sentence parallel (bitext) corpus to learn the mapping from the source to the target language in the scene graph encoding space and decode sentences in the target language,
and \Nii a cross-modal unsupervised feature mapping, which seeks to map the encoded scene graph features from image modality to language modality. We verify the effectiveness of our proposed method on the Chinese image caption generation task. The comparisons against several existing methods demonstrate the effectiveness of our approach.
\end{abstract}

\section{Introduction}

Image captioning has attracted a lot of attention in recent years due to its emerging applications, including assisting visually impaired people, image indexing, virtual assistants, etc. Despite the impressive results achieved by the existing captioning techniques, most of them focus on English because of the availability of image-caption paired datasets,
which can not generalize for languages where such paired dataset is not available. In reality, there are more than 7,100 different languages spoken by billions of people worldwide 
(source: Ethnologue\shortcite{ethno2019}). Building visual-language technologies only for English would deprive a significantly large population of non-English speakers of AI benefits and also leads to ethical concerns,  such as unequal access to resources. Therefore, similar to other NLP tasks (\eg\ parsing, question answering) \cite{hu2020xtreme,Conneau2018xnli,gu2020self}, visual-language tasks should also be extended to multiple languages. However, creating paired captioning datasets for each target language is infeasible, since the labeling process is very time consuming and requires excessive human labor.

To alleviate the aforementioned problem, there have been several attempts in relaxing the requirement of image-caption paired data in the target language~\cite{gu2018unpaired,song2019unpaired}, which rely on paired image-caption data in a pivot language to generate captions in the target language via sentence-level translation. However, even for English, the existing captioning datasets (e.g., MS-COCO~\cite{lin2014microsoft}) are not sufficiently large and comprise only limited object categories, making it challenging to generalize the trained captioners to scenarios in the wild \cite{tran2016rich}. In addition, sentence-level translation relies purely on the text description and can not observe the entire image, which may ignore important contextual semantics and lead to inaccurate translation. Thus, such pivot-based methods fail to fully address the problem. 

Recently, a few works explore image captioning task in an unpaired setting~\cite{feng2019unsupervised,gu2019unpaired,laina2019towards}. Nevertheless, these methods still rely on \jh{manually labeled caption corpus.} For example, \citet{gu2019unpaired}  train their model based on shuffled image-caption pairs of MS-COCO; \citet{feng2019unsupervised}  use an image descriptions corpus from Shutterstock;  Lania et al. (\citeyear{laina2019towards})
create training dataset by sampling the 
images and captions from different image-caption datasets. Despite they belong to \textit{unpaired} methods \textit{in spirit}, 
one could still argue that they depend heavily on the collected caption corpus to get a reasonable cross-modal mapping between vision and language distributions -- a resource that is not always practical to assume. It therefore remains questionable how these methods would perform when there is no caption data at all. To the best of our knowledge, there is yet no work that investigates image captioning without relying on any caption corpus.

Despite the giant gap between images and texts, they are essentially different mediums to describe the same entities and how they are related in the objective world. Such internal logic is the most essential information carried by the medium, which can be leveraged as the bridge to connect data in different modalities.
Scene graph\cite{wang2018scene}, a structural representation that contains 1) the objects, 2) their respective attributes and 3) how they are related as described by the medium (image or text), which has been developed into a mature technique for visual understanding tasks in recent years\cite{johnson2018image}. 
Previous researches on scene graph generation have demonstrated its effectiveness in aiding cross-modal alignment~\cite{yang2019auto, gu2019unpaired}.
{However}, existing scene graph generators are only available in English, which poses challenges for extending its application on other languages. One naive approach is to perform cross-lingual alignment to other target languages by conducting a superficial word-to-word translation on the scene graphs nodes, which neglects the contextual information of the sentence or the image. Since a word on a single node can carry drastically different meanings in various contexts, such an approach often leads to sub-optimal cross-lingual mapping. To address this issue, we propose a novel {Cross-lingual Hierarchical Graph Mapping (HGM)} to effectively conduct the alignment between languages in the scene graph encoding space, which benefits from contextual information by gathering semantics across different levels of the scene graph. Notably, the scene graph translation process can be enhanced by the large-scale parallel corpus (bi-text), which is easily accessible for many languages \cite{espla-etal-2019-paracrawl}.

In this paper, we propose \textbf{UN}pa\textbf{I}red cros\textbf{S}-lingual image capti\textbf{ON}ing (\textbf{UNISON}), \jh{a novel}
approach to generate image captions in the target language without relying on any caption corpus. Our UNISON framework consists of two phases: \Ni a {cross-lingual auto-encoding process} and \Nii a {cross-modal unsupervised feature mapping} (Fig.~\ref{fig:flow_chart}). Using the parallel corpus, 
the {cross-lingual auto-encoding process aims to} \jh{train the HGM} to map a scene graph derived from the source language (English) sentence to the space of the target language (Chinese), and learns to generate a sentence in the target language based on the mapped scene graph.
Then, a {cross-modal feature mapping} (CMM) function is learned in an unsupervised manner, which aligns the image scene graph features from image modality to language modality. \jh{The features in language modality is subsequently mapped by HGM, and then fed to the decoder in phase \Ni to
generate image captions in the target language.}
Our experiments show 
1) the effectiveness of the proposed \textbf{HGM} when conducting cross-lingual alignment(\S5.2) in the scene graph encoding space and 2) the superior performance of our UNISON framework as a whole(\S5.1).

\section{Related Work}
{\textbf{Paired Image Captioning.}}
Previous studies on supervised image captioning mostly follow the popular encoder-decoder framework~\cite{vinyals2015show,rennie2017self,anderson2018bottom}, which mostly focus on generating captions in English since the neural image captioning models require large-scale data of annotated image-caption pairs to achieve good performance. 
To relax the requirement of human effort in caption annotation, \citet{lan2017fluency} propose a fluency-guided learning framework to generate Chinese captions based on pseudo captions, which are translated from English captions.
\citet{yang2019auto} adopt the scene graph as the structured representation to connect image-text domains and generate captions.
\citet{DBLP:conf/eccv/ZhongWC0L20} propose a method to select the important sub-graphs of scene graphs to generate comprehensive captioning.
\citet{Nguyen_2021_ICCV} further close the semantic gap between image and text scene graphs by Human-Object-Interaction labels.

\textbf{Unpaired Image Captioning.}
The main challenge in unpaired image captioning is to learn the captioner without any image-caption pairs. \citet{gu2018unpaired} first propose an approach based on pivot language. They obviate the requirement of paired image-caption data in the target language but still rely on paired image-caption data in the pivot language. \citet{feng2019unsupervised} use a concept-to-sentence model to generate pseudo-image-caption pairs, and align image features and text features in an adversarial manner.
\citet{song2019unpaired} introduce a self-supervised reward to train the pivot-based captioning model on pseudo image-caption pairs.
\citet{gu2019unpaired} propose a scene graph-based method for unpaired image captioning on disordered images and captions.

{\textbf{Summary.}}
While several attempts have been made towards unpaired image captioning, they require \jh{caption corpus to learn a reasonable cross-modal mapping between vision and language distributions}, \eg\  the corpus in~\citep{feng2019unsupervised} is collected from Shutterstock image descriptions, \citet{gu2019unpaired} use the MSCOCO corpus after shuffling the image-caption pairs. Thus, arguably these approaches are not entirely ``unpaired'' \jh{as they rely on the labelled corpus}, limiting their applicability to different languages. Meanwhile, our method generates captions in target language without relying on any caption corpus.

\section{Methods}\label{sec:Methods}

\begin{figure*}[ht!]
	\centering
    \vspace{-2mm}
	\includegraphics[width=1.0\textwidth]{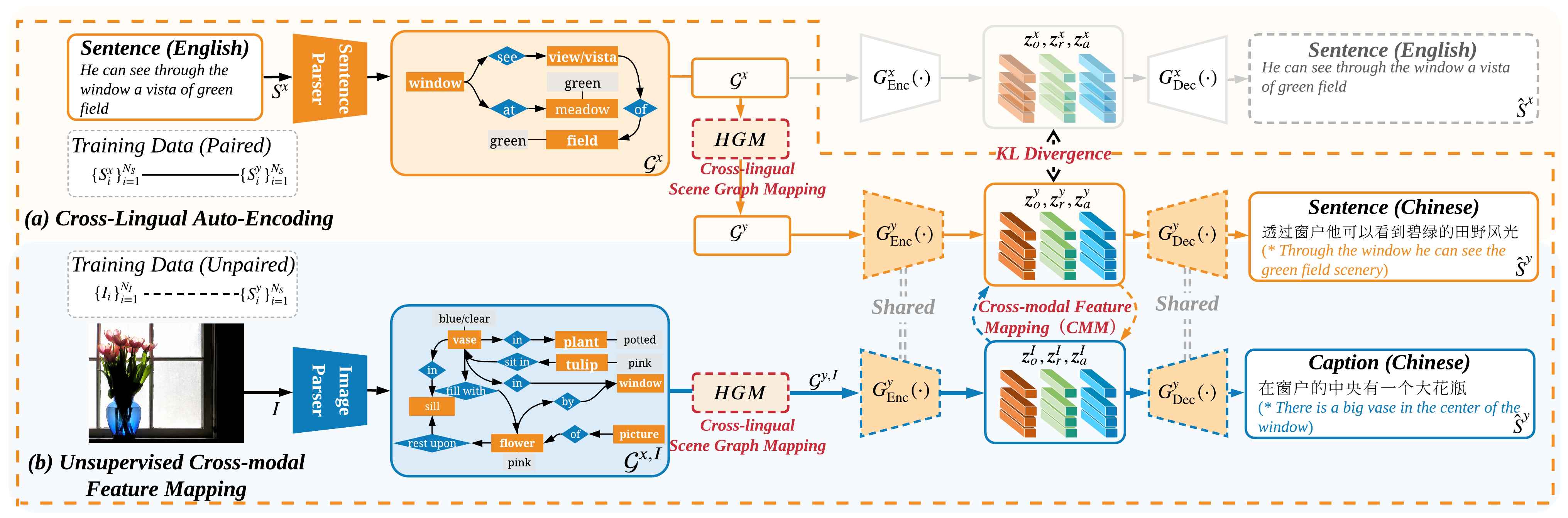}
    \vspace{-5mm}
    \caption{Overview of our \jx{UNISON framework}. It \jx{has} two phases: \jx{\textit{cross-lingual auto-encoding process}} and \jx{\textit{unsupervised cross-modal feature mapping}}. The {cross-lingual} scene graph mapping in the first phase (Top) is designed to map the scene graph from the source language (\eg\ English) to the target language (\eg\ Chinese) without
    {relying on} a scene graph parser in the target language. The unsupervised {cross-modal} feature mapping in the second phase (Bottom) is designed to align the visual modality to textual modality. We mark the object, relationship, and attribute nodes yellow, blue, and grey in the scene graph. The English sentences (marked in gray) in parentheses are translated by google translator for better understanding.}
    \vspace{-3mm}
	\label{fig:flow_chart}
\end{figure*}

\subsection{Preliminary and Our Setting}
In the conventional paired paradigm, image captioning aims to learn a captioner which can generate an image caption $\hat{S}$ for a given image $I$, such that $\hat{S}$ is similar to the ground-truth (GT) caption. Given the image-caption pairs $\{I_i,S_i\}_{i=1}^{N_I}$ , the popular encoder-decoder framework is formulated as:
\begin{equation}
 \mathcal{I}\rightarrow \mathcal{S} : \quad  
 \ I \rightarrow \boldsymbol{v} \rightarrow \hat{S}
 \label{equ:equ_ende}
\end{equation}
{where $\boldsymbol{v}$ denotes the encoded image feature. The
training objective for Eq.~\ref{equ:equ_ende} is to maximize the probability of words in the GT caption given the previous GT words and the image.}

\prj{Compared with paired setting, which relies on paired image-caption data and can not generalize beyond the language used to label the caption, our unpaired setting does not depend on any image-caption pairs and can be extended to other target languages. Specifically, we assume that we have an image dataset $\{I_i\}_{i=1}^{N_I}$ and a source-target parallel corpus dataset $\{(S^x_i,S^y_i)\}_{i=1}^{N_S}$. Our goal is to generate caption $\hat{S}^y$ in the target language $y$ (Chinese) for an image $I$ with the help of unpaired images and parallel corpus. 
}

\subsection{Overall Framework} 
As shown in Fig.~\ref{fig:flow_chart}, there are two phases in our framework: \Ni {a \textbf{cross-lingual auto-encoding process}}
and \Nii a \textbf{cross-modal unsupervised feature mapping}, which can be formulated as the following equations, respectively:
\begin{align}
&{\mathcal{S}^x \rightarrow \mathcal{S}^y: S^x \rightarrow \mathcal{G}^x \Rightarrow \mathcal{G}^y \rightarrow \boldsymbol{z}^y \rightarrow \hat{S}^y\label{eq:equ_ende_2}}\\
&{  \mathcal{I} \rightarrow \mathcal{S}^y \hspace{1.6mm}:  I \rightarrow \mathcal{G}^{x,I} \Rightarrow \mathcal{G}^{y,I} \rightarrow \boldsymbol{z}^{y,I}  \Rightarrow \boldsymbol{z}^{y} \rightarrow \hat{S}^y\label{eq:equ_ende_3}}
\end{align}
where $l \in \left\{x,y\right\}$ is the source or target language; ${\mathcal{S}^l}$ is the sentence in laugnage $l$; $\mathcal{G}^l$ and {$\mathcal{G}^{l,I}$} are the scene graphs for the language $l$ in sentence modality and image modality ($I$), respectively; $\boldsymbol{z}^l$ and {$\boldsymbol{z}^{l,I}$} are the encoded scene graph features for $\mathcal{G}^l$ and {$\mathcal{G}^{l,I}$}, respectively.

The \textbf{cross-lingual auto-encoding process} (shown in top of Fig.~\ref{fig:flow_chart}) aims to generate a sentence in the target language given a scene graph in the source language: we first extract the sentence scene graph $\mathcal{G}^x$ from each (English) sentence $S^x$ using a sentence scene graph parser, and map it to $\mathcal{G}^y$ via our proposed HGM (detail in later section). 
Then we feed
$\mathcal{G}^y$ to 
the encoder to produce the scene graph features
$\boldsymbol{z}^y$, \prj{which the decoder then takes as inputs to generate
$\hat{S}^y$}. Note that the mapping from $\mathcal{G}^x$ to $\mathcal{G}^y$ is done at the embedding level, \ie\ no symbolic $\mathcal{G}^y$ is constructed. This phase addresses the misalignment among different language domains.

The \textbf{cross-modal unsupervised feature mapping} (shown in the bottom part of Fig.~\ref{fig:flow_chart}) closes the gap between image modality and language modality: we first extract the image scene graph $\mathcal{G}^{x,I}$ from image $I$, which is in source language $x$ (English). After that, we map the $\mathcal{G}^{x,I}$ to $\mathcal{G}^{y,I}$ with the HGM (shared
with the first phase).
As shown in Eq.~\ref{eq:equ_ende_3}, a cross-modal mapping function ($ \boldsymbol{z}^{y,I}   \Rightarrow  \boldsymbol{z}^{y}$) is learned, which maps the encoded image scene graph features from {image modality to language modality}. Once mapped to  $\boldsymbol{z}^{y}$, we can use the sentence decoder to generate $\hat{S}^y$. We further elaborate each phase in detail below.

\subsection{Cross-lingual Auto-encoding Process}
\paragraph{Scene Graph.}
A scene graph $\mathcal{G}=(\mathcal{V},\mathcal{E})$ contains three kinds of nodes: object, relationship and attribute nodes.
Let object $o_i$ denote the $i$-th object. The triplet $\langle o_i, r_{i,j}, o_j \rangle$ in $\mathcal{G}$ is composed of two objects: $o_i$ (as subject role) and $o_j$ (as object role), along with their relation $r_{i,j}$. As each object may have a set of attributes, we denote $a_i^k$ as the $k$-th attribute of object $o_i$. { To generate an image scene graph $\mathcal{G}^I$, we build the image scene graph generator based on Faster-RCNN~\cite{ren2015faster} and MOTIFS~\cite{zellers2018neural}.} To generate sentence scene graph $\mathcal{G}^x$, we first convert each sentence into a dependency tree with a syntactic parser~\cite{anderson2016spice}, and then apply a rule-based method~\cite{schuster2015generating} to build the graph. The $\mathcal{G}^y$ is mapped from $\mathcal{G}^x$ through our HGM module.

\paragraph{Cross-lingual Hierarchical Graph Mapping (HGM).}

Our hierarchical graph mapping contains three levels: \Ni word-level mapping, \Nii sub-graph mapping, and \Niii full-graph mapping. The semantic information from all three levels are fused in an self-adaptive manner via a self-gated mechanism, which effectively takes into account the structures and relations from the context. 

\begin{figure}[t!]
	\centering
	\includegraphics[width=0.48\textwidth]{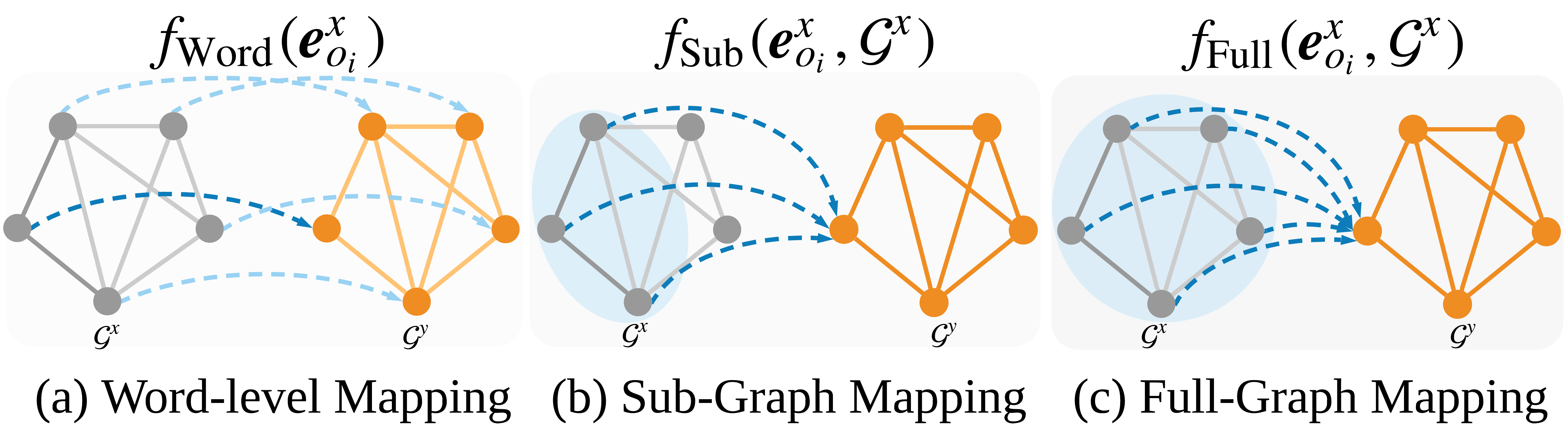}
    \vspace{-7mm}
	\caption{Illustration of our hierarchical scene graph mapping. Sub-graph mapping only considers those directly connected nodes, while full-graph mapping considers all the nodes in the scene graph.}
	\label{fig:hgm}
    \vspace{-7mm}
\end{figure}

The {proposed} HGM is illustrated in Fig.~\ref{fig:hgm}. Let $\langle \boldsymbol{e}_{o_i}^l, \boldsymbol{e}_{r_{i,j}}^l, \boldsymbol{e}_{o_j}^l\rangle \in \mathcal{G}^l$ denote the triplet for relation $r_{i,j}^l$ in language $l$, where $\boldsymbol{e}_{o_i}^l$, $\boldsymbol{e}_{o_j}^l$ and $\boldsymbol{e}_{r_{i,j}}^l$ are the embeddings representing subject $o_i^l$, object $o_i^l$, and relationship $r_{i,j}^l$. Formally, our hierarchical graph mapping from language $x$ to language $y$ can be expressed as:
\begin{align}
    \langle \boldsymbol{e}_{o_i}^y, \boldsymbol{e}_{r_{i,j}}^y,\boldsymbol{e}_{o_j}^y \rangle =& \langle f_{\text{HGM}}(\boldsymbol{e}_{o_i}^x,\mathcal{G}^x), f_{\text{Word}}(\boldsymbol{e}_{r_{i,j}}^x),
    \nonumber \\
    & f_{\text{HGM}}(\boldsymbol{e}_{o_j}^x,\mathcal{G}^x)\rangle\label{eq:mapping}
    \\
    f_{\text{HGM}}(\boldsymbol{e}_{o_i}^x,\mathcal{G}^x) = &\alpha_w f_{\text{Word}}(\boldsymbol{e}_{o_i}^x)+ \alpha_s f_{\text{Sub}}(\boldsymbol{e}_{o_i}^x,\mathcal{G}^x) \nonumber\\
    & +\alpha_f f_{\text{Full}}(\boldsymbol{e}_{o_i}^x,\mathcal{G}^x) \label{eq:hgm}\\
    \langle \alpha_w,\alpha_s,\alpha_f \rangle= &\text{softmax} \big( f_{{\text{MLP}}} (f_{\text{Word}}(\boldsymbol{e}_{o_i}^x))\big)
    \label{eq:self_gate}
\end{align}
where $\boldsymbol{e}_{o_i}^y$, $\boldsymbol{e}_{o_j}^y$ and $\boldsymbol{e}_{r_{i,j}}^y$ are the mapped embeddings in target language $y$; 
$\alpha_w,\alpha_s, \text{ and } \alpha_f$ are the level-wise importance weights calculated by
Eq.~\ref{eq:self_gate};   $f_{{\text{MLP}}}(\cdot)$ {represents a multi-layer perception (MLP) composed of three fully-connected (FC) layers with ReLU activations}.

\textit{Word-level Mapping.}
The word-level mapping relies on a retrieval function $f_{\text{Word}}(.)$: after obtaining an embedding in language $x$, $f_{\text{Word}}(.)$ retrieves the most similar embedding in language $y$ from a cross-lingual word embedding space as illustrated in Fig.~\ref{fig:hgm}(a)\prj{, where cosine similarity is used to measure the distance.} In practice, we adopt the pre-trained common space trained on Wikipedia following ~\cite{joulin2018loss}. 
The retrieved embedding is then passed to an FC layer to obtain a high-dimension embedding in language $y$.

\textit{Graph-level Mapping.}
Since the relation and structure of the surrounding nodes encode crucial context information,  \jh{ we also introduce the node mapping} \jx{with} graph-level information (as illustrated by Fig.~\ref{fig:hgm}(b) and \ref{fig:hgm}(c)): \jh{namely} {sub-graph mapping} ($f_{\text{Sub}}$) and {full-graph mapping} ($f_{\text{Full}}$),
\jh{which first construct the contextualized embedding in graph-level and then conduct the cross-lingual mapping on the produced embedding. More specifically,} for sub-graph mapping,
the contextualized embedding is computed by: $\sum_{k=1}^{N'_{o}}
sconv(\boldsymbol{e}_{o_i}^x,\boldsymbol{e}_{o_k}
^x) / N'_{o}$, where $N'_{o}$ is the total number of nodes directly connected to node $o_i$, and 
$sconv(\cdot)$ is the spatial convolution operation~\cite{yang2019auto}.  For full-graph mapping,
the contextualized embedding is calculated by an attention module: $\sum_{k=1}^{N_o} \alpha_k \boldsymbol{e}_{o_k}^x$, where $\alpha_k$ is calculated by softmax over all the object embeddings $\boldsymbol{e}_{o_{1:N_o}}^x$. 
Both $f_{\text{Sub}}$ and $f_{\text{Full}}$
use a linear mapping to project the resulted contextualized (English) embedding to the target (Chinese) embedding space.
We consider graph-level mapping only for the object nodes since relationships only exist between objects. For relationship and attribute nodes, only word-level mapping is performed.

\textit{Self-gated Adaptive Fusion.}
\jh{To leverage the complementary advantages of information in different levels, we propose a self-gate mechanism to adaptively adjust the importance weights when fusing the embeddings.}
Specifically, the importance scores are calculated \jh{based on} the word-level embeddings by passing it through a three-class MLP and a softmax function (Eq.~\ref{eq:self_gate}). \jh{Compared with directly concatenating the embeddings from different levels, which assigns them with equal importance, our fusing mechanism adaptively concentrates on important information and suppress the noises when the context becomes sophisticated.}

\paragraph{Scene Graph Encoder.}
We encode the $\mathcal{G}^x$ and $\mathcal{G}^y$(mapped by the HGM) with two scene graph encoders $G_{\text{Enc}}^x (\cdot)$ and $G_{\text{Enc}}^y (\cdot)$, which are  implemented by spatial graph convolutions.
The output of each scene graph encoder can be formulated as:
\begin{equation}
    \boldsymbol{f}_{o_{1:N_o^l}}^l, \boldsymbol{f}_{r_{1:N_r^l}}^l,\boldsymbol{f}_{a_{1:N_a^l}}^l = G_{\text{Enc}}^l(\mathcal{G}^l) , \quad l \in \left\{x,y\right\}
\end{equation}
where $\boldsymbol{f}_{o_{1:N_o^l}}^l$, $\boldsymbol{f}_{r_{1:N_r^l}}^l$, and $\boldsymbol{f}_{a_{1:N_a^l}}^l$ denote the set of encoded object embeddings, relationship embeddings, and attribute embeddings, respectively. Each object embedding $\boldsymbol{f}_{o_{i}}^l$ is calculated by considering relationship triplets $\langle \boldsymbol{e}_{\text{sub}(o_i)}^l,  \boldsymbol{e}_{r_{\text{sub}(o_i),i}}^l, \boldsymbol{e}_{o_i}^l \rangle$ and $\langle \boldsymbol{e}_{o_j}^l, \boldsymbol{e}_{r_{j, \text{obj}(o_i)}}^l, \boldsymbol{e}_{\text{obj}(o_i)}^l \rangle$;
$\text{sub}(o_i)$ represents the subjects where $o_i$ acts as an object, and $\text{obj}(o_i)$ represents the objects where $o_i$ plays the subject role. $\boldsymbol{f}_{r_{i}}^l$ is calculated based on relationship triplet $\langle \boldsymbol{e}_{o_i}^l, \boldsymbol{e}_{r_{i,j}}^l, \boldsymbol{e}_{o_j}^l \rangle$. $\boldsymbol{f}_{a_{i}}^l$ is the attribute embedding calculated by object $o_i$ and its associated attributes.

\paragraph{Sentence Decoder.}
As shown in Fig.~\ref{fig:flow_chart}, we have two decoders: $G_{\text{Dec}}^x (\cdot)$ and $G_{\text{Dec}}^y (\cdot)$. Each decoder is composed of three attention modules and an LSTM-based decoder. {It takes the encoded scene graph features as input and generates the captions. The decoding process is defined as:}
\begin{align}
    \boldsymbol{o}_t^l, \boldsymbol{h}_t^l &=\text{G}_{\text{Dec}}^l \Big(f_{\text{Triplet}}\big([\boldsymbol{z}_{o}^l,\boldsymbol{z}_{r}^l,\boldsymbol{z}_{a}^l]\big),\boldsymbol{h}_{t-1}^l,\hat{s}_{t-1}^l\Big) \label{eq:rnn_dec} \\
    \hat{s}_t^l & \sim \text{softmax}(\boldsymbol{W}_o\boldsymbol{o}_t^l)\label{eq:soft_out}
\end{align}
\hspace{0mm}where $l \in \left\{x,y\right\}$, $\hat{s}_t^l$ is the $t$-th decoded word drawn from the dictionary according to the softmax probability, $\boldsymbol{W}_o$ is a learnable weight matrix, $\boldsymbol{o}_t^l$ is the cell output of the decoder, $\boldsymbol{h}_t^l$ is the hidden state.  $f_{\text{Triplet}}(\cdot)$ is a non-linear mapping function that takes the concatenated features {as} input and outputs the triplet level feature.
$\boldsymbol{z}_{o_i}^l$ is calculated by the attention module defined as: $\sum _i^{N_o^l} \alpha_{o_i}^l\boldsymbol{f}_{o_i}^l$, {where $\alpha_{o_i}^l$ is the attention weight calculated by the softmax operation over $\boldsymbol{f}_{o_{1:N_o^l}}^l$.}
$\boldsymbol{z}_{r_i}^l$ and $\boldsymbol{z}_{a_i}^l$ are calculated in a similar way.

\vspace{-0.4em}
\subsubsection{Joint-training Mechanism.}\label{sec:learning}
Inspired by the fact that common structures exist in the encoded scene graph space that are language-agnostic, which may be leveraged to benefit the encoding process, we propose joint training mechanism to enhance the features in target language with the help of features in the source language. In practice, we train a separate scene graph encoder for each language in parallel, then align the encoded scene graph features by enforcing them to be semantically close.

Specifically, we train the scene graph encoders ($G_{\text{Enc}}^x$ and $G_{\text{Enc}}^y$), sentence decoders ($G_{\text{Dec}}^x$ and $G_{\text{Dec}}^y$), and the cross-lingual {HGM} module ($\mathcal{G}^x \Rightarrow \mathcal{G}^y  $), supervised by a parallel corpus.
The two graph encoders encode $\mathcal{G}^x$ and $\mathcal{G}^y$ into feature representations and predict sentences ($\hat{S}^x$ and $\hat{S}^y$) with the decoders. We minimize the following loss:
\begin{equation}
    \begin{split}
    \mathcal{L}_{\text{XE}} =&-\sum_t \log P_{\theta_{\mathcal{G}^x \rightarrow S^x}}(s^x_t|s^x_{0:t-1},\mathcal{G}^x)\\
    &-\sum_t \log P_{\theta_{\mathcal{G}^y \rightarrow S^y}}(s^y_t|s^y_{0:t-1},\mathcal{G}^y
    )
    \end{split}\label{eq:xe_cross_lingual}   
\end{equation}
where the $s^x_t$ and $s^y_t$ are the ground truth words, $\mathcal{G}^x$ and $\mathcal{G}^y$ are the sentence scene graphs in different languages with $\mathcal{G}^y$ being derived from $\mathcal{G}^x$ using our HGM, 
$\theta_{\mathcal{G}^x \rightarrow S^x}$ and $\theta_{\mathcal{G}^y \rightarrow S^y}$ are the parameters of two encoder-decoder models.

To close the semantic gap between the encoded scene graph features $\{ \boldsymbol{z}_{o}^x, \boldsymbol{z}_{a}^x, \boldsymbol{z}_{r}^x\}$ and $\{ \boldsymbol{z}_{o}^y, \boldsymbol{z}_{a}^y, \boldsymbol{z}_{r}^y\}$, we introduce a Kullback–Leibler (KL) divergence loss:
\begin{align}
       \mathcal{L}_{\text{KL}} 
        =&\exp\big(\text{KL}(p(\boldsymbol{z}_o^x)||p(\boldsymbol{z}_o^y)\big) +\exp\big(\text{KL}(p(\boldsymbol{z}_a^x)||p(\boldsymbol{z}_a^y)\big)\nonumber\\
        &+\exp\big(\text{KL}(p(\boldsymbol{z}_r^x)||p(\boldsymbol{z}_r^y)\big)\label{eq:kd_cross_lingual} 
\end{align}
where $p(\cdot)$ is composed of a linear layer that maps the input features to a low-dimension $d_c$, followed by a softmax to get a probability distribution.
The overall objective of our joint training mechanism is as follows: $\mathcal{L}_{\text{Phase 1}}=\mathcal{L}_{\text{XE}} +\mathcal{L}_{\text{KL}} $.

\subsection{Unsupervised Cross-modal Feature Mapping}
To adapt the learned model from sentence modality to image modality, 
we drew inspiration from ~\cite{gu2019unpaired} and adopt CycleGAN~\cite{zhu2017unpaired} to align the features. For each type {$p\in \left\{o,r,a\right\}$} of triplet embedding in Eq.~\ref{eq:rnn_dec}, we have two mapping functions: $g_{I\rightarrow y}^p(\cdot)$ and $g_{y\rightarrow I}^p(\cdot)$, where $g_{I\rightarrow y}^p(\cdot)$ maps the features from image modality to the sentence modality, and $g_{y \rightarrow I}^p(\cdot)$ maps from sentence modality to the image modality. Note that we freeze the cross-lingual mapping module trained in the first phase. The training objective for cross-modal feature mapping is defined as:
\begin{equation}
    \mathcal{L}_{\text{CycleGAN}}^p = \mathcal{L}_{\text{GAN}}^{I\rightarrow y}+\mathcal{L}_{\text{GAN}}^{y\rightarrow I}+\lambda \mathcal{L}_{\text{cyc}}^{I \leftrightarrow y}
    \label{eq:cycle_gan_final}
\end{equation}
where $\mathcal{L}_{\text{cyc}}^{I \leftrightarrow y}$ is a cycle consistency loss, $\mathcal{L}_{\text{GAN}}^{I\rightarrow y}$ and {$\mathcal{L}_{\text{GAN}}^{y\rightarrow I}$ are the adversarial losses for the mapping functions with respect to the discriminators}.

Specifically, the objective of the mapping function $g_{I\rightarrow y}^p(\cdot)$
is to fool the discriminator $D_y^p$ through adversarial learning. \prj{We formulate the objective function for cross-modal mapping as:}
\begin{equation}
    \small
    \mathcal{L}_{\text{GAN}}^{I\rightarrow y} =  \mathbb{E}_{S}[\log D_y^p(\boldsymbol{z}_p^y)]  + \mathbb{E}_{I}[\log (1-D_y^p(g_{I \rightarrow y}^p(\boldsymbol{z}_p^I))] \label{eq:gan_i_s}
\end{equation}
where $\boldsymbol{z}_p^y$ and $\boldsymbol{z}_p^I$ are the encoded  embeddings for sentence scene graph $\mathcal{G}^y$ and image scene graph $\mathcal{G}^{y,I}$, respectively. The adversarial loss for sentence to image mapping $\mathcal{L}_{\text{GAN}}^{y\rightarrow I}$ is similarly defined. 
The cycle consistency loss $\mathcal{L}_{\text{cyc}}^{I \leftrightarrow y}$ is designed to regularize the training and make the mapping functions cycle-consistent:
\begin{align}
    \mathcal{L}_{\text{cyc}}^{I \leftrightarrow y} = &\mathbb{E}_{I}[\| g_{S\rightarrow I}^p\big(g_{I\rightarrow S}^p(\boldsymbol{z}_p^I)\big)-\boldsymbol{z}_p^I \|_1]\nonumber\\
    &+\mathbb{E}_{y}[\| g_{I\rightarrow y}^p\big(g_{y\rightarrow I}^p(\boldsymbol{z}_p^y)\big)-\boldsymbol{z}_p^y \|_1]\label{eq:gan_cyc}
\end{align}
The overall training objective for phase 2 becomes:  $\mathcal{L}_{\text{Phase 2}}=\mathcal{L}_{\text{CycleGAN}}^o+\mathcal{L}_{\text{CycleGAN}}^a+\mathcal{L}_{\text{CycleGAN}}^r$.

\subsection{Inference of the UNISON Framework}
During inference, given an image $I$, we first extract the image scene graph $\mathcal{G}^{x,I}$ with a pre-trained image scene graph generator and then map the $\mathcal{G}^{x,I}$ in $x$ (English) to $\mathcal{G}^{y,I}$ in $y$ (Chinese) with our HGM module. After that, we encode $\mathcal{G}^{y,I}$ with $G_{\text{Enc}}^y(\cdot)$ and map the encoded {features} to the language domain through $g^p_{I\rightarrow y}(\cdot)$. The mapped features are then fed to the LSTM-based sentence decoder $G_{\text{Dec}}^y(\cdot)$ to generate the image caption $\hat{S}^y$ in target language $y$.

\section{Experiments}
\subsection{Datasets and Setting}\label{sec:dataset_setting}
\paragraph{Datasets.}
In this paper, we use English and Chinese as the source and target languages, respectively.
For cross-lingual auto-encoding, we collect a paired English-Chinese corpus from existing MT datasets, including WMT19~\cite{barrault2019findings}, AIC\_MT~\cite{wu2017ai}, UM~\cite{tian-etal-2014-um}, and Trans-zh ~\cite{bright_xu_2019_3402023}. 
We filter the sentences in MT datasets according to an {existing caption-style} dictionary {containing} {7,096} words \jh{in} \citet{li2019coco}.
For the first phase, we use 151,613 sentence pairs for training, 5,000 sentence pairs for validation, and 5,000 pairs for testing.
For the second phase, following \citet{li2019coco}, we use 18,341 {training} images from MS-COCO and randomly select 18,341 Chinese sentences from the training split of the MT corpus.
{During evaluation, we use the validation and testing splits in COCO-CN.}
More details are in the appendix.

\begin{table}[ht]
\vspace{-2mm}
\begin{center}
\setlength{\tabcolsep}{5pt}{
\begin{tabular}{l|c|c|c|c}
\hline
Corpus &  0 Obj/$\mathcal{G}$ & 1 Obj/$\mathcal{G}$ &  2 Obj/$\mathcal{G}$ & $\geqslant$3 Obj/$\mathcal{G}$\\
\hline
Raw  & 17.7\% & 42.6\%  & 24.4\% &  15.4\% \\
Back-Trans. & 12.3\%  & 13.3\%  & 15.1\% & 59.3\% \\
\hline
\end{tabular}
}
\end{center}
\vspace{-2mm}
\caption{{Statistics of the English sentence scene graphs, where n Obj/$\mathcal{G}$ denotes the number of object in a scene graph, $\geqslant$ means greater than or equal to 3.}}\label{sg_stat}
\vspace{-4mm}
\end{table}

\begin{table*}[h]
\small
\begin{center}
\renewcommand\arraystretch{1.1}{
\setlength{\tabcolsep}{3pt}{
\begin{tabular}{l|l|c|c|c|c|c|c|c}
\hline
& Method  & B@1 & B@2 & B@3 & B@4 & METEOR & ROUGH & CIDEr \\
\hline
\multicolumn{9}{l}{\textit{{Setting w/o caption corpus}}}\\
\hline
\parbox[t]{4mm}{\multirow{2}{*}{\rotatebox[origin=c]{90}{\small{Un.}}}}
& Graph-Aligh(En)\cite{gu2019unpaired} +GoogleTrans. & 39.2 & 16.7&6.5 &2.3 &13.2 &26.5 &9.3 \\ 
& UNISON & \textbf{44.9}  & \textbf{19.9} & \textbf{8.6} &  \textbf{3.3}  &  \textbf{16.5} & \textbf{29.6} & \textbf{12.7} \\ 
\hline
\hline
\multicolumn{9}{l}{\textit{{Setting w/ caption corpus}}}\\
\hline
\parbox[t]{4mm}{\multirow{2}{*}{\rotatebox[origin=c]{90}{\small{Pair}}}}
& FC-2k (En)\cite{rennie2017self}+GoogleTrans. & 58.9 &38.0 & 23.5 &  14.3 &23.5 &40.2 & 47.3 \\
& FC-2k (Cn, Pseudo COCO)\cite{rennie2017self}
& 60.4 & 40.7 & 26.8 & 17.3 & 24.0 & 43.6 & 52.7 \\
\hline
\parbox[t]{4mm}{\multirow{1}{*}{\rotatebox[origin=c]{90}{\scriptsize{Un.}}}}
& UNISON & \textbf{63.4} & \textbf{43.2} & \textbf{29.5} & \textbf{17.9} &  \textbf{24.5} & \textbf{45.1} &\textbf{53.5} \\
\hline
\end{tabular}}}
\end{center}
\vspace{-4mm}
\caption{ Performance comparisons on the test split of COCO-CN. `Un.' is short for Unpaired. B@$n$ is short for BLEU-$n$. `En' and `Cn' in the parentheses represent English and Chinese, respectively. `GoogleTrans' stands for google translator.}\label{tab:overall}
\vspace{-3mm}
\end{table*}

\paragraph{Preprocessing.}
{We extract the image scene graph with MOTIFS~\cite{zellers2018neural} pretrained on VG~\cite{krishna2017visual}.}
We tokenize and lowercase the English sentences, then replace the tokens appeared less than five times with \textsc{UNK}, resulting in a vocabulary size of 13,194. We segment the Chinese sentences with \textit{Jieba}\footnote{https://github.com/fxsjy/jieba} and apply the same treatment for words appeared less than five times, resulting in a vocabulary size of 11,731.
The English sentence scene graphs are extracted with the parser proposed by \cite{anderson2016spice}. 
We augment the English sentences with the pre-trained back-translators~\cite{ng2019facebook}, resulting in
808,065 English sentences in total, which helps enrich the English sentence scene graphs. Specifically, the statistics in Table~\ref{sg_stat} shows that
\jh{
the percentage of scene graphs containing more than 3 objects is increased from 15.4\% to 59.3\%.}
\jh{Quantitative and qualitative results in Appendix further shows that data augmentation effectively enrich the scene graphs and benefit the sentence generation.}

\subsection{Implementation Details}
During cross-lingual auto-encoding phase, we set the dimension of scene graph embeddings to 1,000 and $d_c$ to 100. 
\prj{LSTM with 2 layers is adopted to construct the decoder, whose hidden size is 1000.}
\prj{ We start by initializing the graph mapping from a pre-trained common space \cite{joulin2018loss} to stabilize training. The cross-lingual encoder-decoder is firstly trained with the $\mathcal{L}_{\text{XE}}$ for 80 epochs, then with joint loss $\mathcal{L}_{\text{Phase 1}}$ for 20 epochs. }
 
During unsupervised cross-modal mapping phase, we learn the cross-modal feature mapping on the unpaired MS-COCO images and translation corpus. Specifically, we \prj{inherit and freeze the} parameters of the Chinese scene graph encoder, HGM, and Chinese sentence decoder from \jh{cross-lingual auto-encoding process}. The cross-modal mapping functions and discriminators are learned with $\mathcal{L}_{\text{Phase 2}}$.
We optimize the model with Adam, batch size of 50, and learning rate of $5 \times 10^{-5}$.
\jh{The discriminators are implemented with a linear layer of dimension 1,000 and a LeakyReLU activation.  We set $\lambda$ to 10.}
During inference, we use beam search with a beam size of 5. 
We use the popular BLEU \cite{papineni2002bleu}, CIDEr \cite{vedantam2015cider}, METEOR \cite{denkowski2014meteor} and ROUGE \cite{lin2004rouge} for evaluation. \prj{More details are attached to the Appendix.}

\subsection{Model Statement}
{To gain insights into the effectiveness of our
\prj{HGM, we construct ablative models by progressively introducing cross-lingual graph-mappings in different levels}:}

\noindent \textbf{$\text{GM}_{\textsc{base}}$} is our baseline model, which adopts Google's MT system~\cite{wu2016google} to symbolically map the scene graph from English to Chinese in a node-to-node manner. 

\noindent \textbf{$\text{GM}_{\textsc{word}}$} maps the English scene graph to Chinese through word-level mapping in the scene graph encoding space.

\noindent \textbf{$\text{GM}_{\textsc{Word+Sub.}}$} considers both word-level and subgraph-level mappings by directly concatenating them.

\noindent \textbf{$\text{HGM}_{\textsc{base}}$} considers mappings across all levels, {which are directly concatenated
and passed through an FC layer.}

\noindent \textbf{HGM} {is similar to HGM-base, except that it adopts a self-gated fusion to adaptively fuse the three features, as illustrated by Eq.~\ref{eq:hgm} and Eq.~\ref{eq:self_gate}.}

\section{Results and Analysis}\label{sec:quantitative}

\subsection{Overall Results}
We demonstrate the superior performance of the proposed UNISON framework on Chinese image captions generation task. We first compare \jx{UNISON} with the SOTA unpaired method Graph-Align\cite{gu2019unpaired} under the setting without using any caption corpus. More specifically, we run the Graph-Align\footnote{Code is acquired from the first author of \cite{gu2019unpaired}.}  and translate the generated English captions to Chinese by google translator for comparison. From the result in Table~\ref{tab:overall}, we can find that our method significantly surpasses Graph-Align with translation, demonstrating that translation in graph level is superior to  translation in sentence level. \jh{This is reasonable since the graph level alignment is able to consider structural and relational information of the whole image, while sentence level translation suffers from information loss as it can only observe the predicted sentences, and can be severely affected if the translation tools perform poorly.} We do not compare with the other unpaired method \cite{song2019unpaired} here, as the dataset and codes are not publicly available.

To further verify the effectiveness of our framework, we compare UNISON with the supervised pipeline methods: 
 \textbf{\Ni}\textit{FC-2k(En)+Trans.} We train the FC-2k model on image-caption pairs(En) of MS-COCO and translate the generated captions(En) to caption(Cn) using Google translator;  \textbf{\Nii}\textit{FC-2k(Pseudo).} We train the FC-2k model on pseudo Chinese image-caption pairs of MS-COCO, where the captions(Cn) are translated by Google translator from captions(En). 
For such comparisons, we fine-tune our cross-lingual mapping on the unpaired captions. The results show that our method significantly and consistently outperforms the \textit{FC-2k(En)+Trans.} and \textit{FC-2k(Pseudo)} models in all metrics, despite our unpaired setting is much weaker. 

\subsection{Effectiveness of Cross-lingual Alignment}
\paragraph{Analyzing the superior performance of HGM.}
We conduct experiments on MT task to demonstrate our HGM's effectiveness in cross-lingual alignment, which is shown in Table~\ref{tab:graph_mapping_methods}. The advantage of HGM lies in four aspects: (1) The cross-lingual graph translation is effective. Our HGM and its variants achieve considerably higher performance compared with $G_{\textsc{EN}}$, which directly generates Chinese sentences based on English scene graphs. (2) The cross-lingual alignment in the encoding space is superior than direct symbolic translation. $\text{GM}_{\textsc{word}}$ achieve considerably higher performance compared with $\text{GM}_{\textsc{base}}$, which
proves that the scene graph encoding contains richer information and is more suitable for cross-lingual alignment. (3) Performing node mapping considering features across different graph levels boosts the performance. \jh{When we consider full-graph and sub-graph level features, the cross-lingual alignment starts achieving significant performance improvement, which verifies the importance of structural and relational information in the context.} E.g., HGM outperforms $\text{GM}_{\textsc{word}}$ in B@1, B@2, B@3, B@4, METEOR, and ROUGE metrics by 8.2\%, 17.9\%, 29.9\%, 38.8\%, 10.4\%, 13.1\%, respectively. (4) The adaptive self-gate fusion mechanism is beneficial. We can observe that $\text{HGM}_{\textsc{base}}$ is surpassed by $\text{HGM}$ by a large margin.
As shown in Table~\ref{joint+cyclegan}, the role of self-gate fusion becomes more essential when HGM is applied to image scene graphs.

\begin{table}[ht]
\small
\begin{center}
    \setlength{\tabcolsep}{1.5pt}{
\begin{tabular}{l|c|c|c|c|c|c|c|c}
\hline
Method & SG-Map  & S-Gate & B@1 &  B@2 & B@3 & B@4 & M & R \\
\hline
$\text{G}_{\textsc{en}}$ & \xmark & \xmark & 25.0 & 13.8 & 8.2 & 5.2 & 14.4 & 27.3\\
$\text{GM}_{\textsc{base}}$ & \cmark& \xmark &  {26.6} & {15.4} & {10.0} & {7.3} & {15.1} & {28.1} \\
\hline
$\text{GM}_{\textsc{word}}$ & \cmark & \xmark & 28.1   &   16.2  &  10.7   & 8.0    &  15.4   &  28.2\\
$\text{GM}_{\textsc{word+sub.}}$ & \cmark & \xmark  & 29.2  &  17.9 &   12.6  &  9.9   &  16.3  & 30.2\\
$\text{HGM}_{\textsc{base}}$ & \cmark & \xmark &   29.6  &  18.1  & 12.8  & 9.9& 16.5 &  30.4 \\
HGM 
& \cmark & \cmark  &  \textbf{30.4}  & \textbf{19.1}  &  \textbf{13.9}  & \textbf{11.1}  & \textbf{17.0} & \textbf{31.9} \\
\hline
\end{tabular}
}
\end{center}
\vspace{-4mm}
\caption{Performance comparison between variants of HGM on Chinese sentence generation task. Test split of MT corpus is used for evaluation. `SG-Map' is cross-lingual scene graphs mapping. `S-Gate' is self-gate fusion mechanism.}\label{tab:graph_mapping_methods}
\vspace{-2mm}
\end{table}

\begin{table}[h]
\small
\vspace{-1mm}
\begin{center}
    \setlength{\tabcolsep}{1.5pt}{
\begin{tabular}{l|c|c|c|c|c|c|c|c}
\hline
Method & $\mathcal{L}_{\text{XE}}$ & $\mathcal{L}_{\text{KL}}$& B@1 &  B@2 & B@3 & B@4 & M & R \\
\hline
$\text{GM}_{\textsc{word}}$ & \cmark &  \cmark & 28.1   &   16.2  &  10.7   & 8.0    &  15.4   &  28.2\\
\quad-w/o joint& \cmark &  \xmark&  -0.4  &   -0.4  &   -0.3  &   -0.3  &  -0.1   &  -0.2 \\
\hline
$\text{GM}_{\textsc{word+sub.}}$ & \cmark &  \cmark & 29.2  &  17.9 &   12.6  &  9.9   &  16.3  & 30.2\\
\quad-w/o joint & \cmark &  \xmark&  -0.3   & -0.3   &  -0.3  &  -0.4 &   -0.2 & -0.1\\
\hline
{HGM }  & \cmark &  \cmark&  \textbf{30.4}  & \textbf{19.1}  &  \textbf{13.9}  & \textbf{11.1}  & \textbf{17.0} & \textbf{31.9} \\
\quad-w/o joint  & \cmark &  \xmark& -0.5  & -0.3 &  -0.3  & -0.3  & -0.2 & -0.4 \\
\hline
\end{tabular}
}
\end{center}
\vspace{-4mm}
\caption{Effectiveness of joint training in cross-lingual auto-encoding. Performance comparison of the Chinese sentence generation on the test split of MT corpus. }\label{tab:graph_mapping}
\vspace{-4mm}
\end{table}

\paragraph{{Joint training benefits the enconding process.}}
We train our models using the joint loss $\mathcal{L}_{\text{Phase 1}}$, where $\mathcal{L}_{\text{KL}}$ enforces the distributions of latent scene graph embeddings {between different languages} to be close. Table \ref{tab:graph_mapping} shows that the models trained with joint loss
{consistently outperforms their counterparts with only $\mathcal{L}_{\text{XE}}$ for all metrics,}
which indicates that the encoding process of the target language can benefit from the source language.

\subsection{Effectiveness of cross-modal feature mapping}
To bridge the gap between image and language modalities, we learn the cross-modal feature mapping
in an unsupervised manner. 

Table~\ref{joint+cyclegan} shows the performance of Chinese image captioners with and without CMM.
We can see that adversarial training can {consistently} improve the model's performance. Specifically, CMM can boost the performance of our HGM by 3.8\%(B@1), 2.8\%(B@2), 1.7\%(B@3), 0.7\%(B@4), 1.2\%(ROUGE), 3.0\%(CIDER), respectively. {Notably, $\text{GM}_{\textsc{Word+Sub.}}$ and $\text{HGM}_{\textsc{base}}$ perform even worse than $\text{GM}_{\textsc{base}}$, which is because the generated image scene graphs are noisy with repeated relation triples (as explained in \S\ref{sec:qualitative}), leading to degradation on contextualized cross-lingual graph mapping (sub-graph and full-graph), whereas self-gated fusion can tackle this problem by decreasing the importance of noisy graph-level mapping.}

\begin{table}[h]
\small
\vspace{-1.5mm}
\begin{center}
\setlength{\tabcolsep}{1.5pt}{
\begin{tabular}{l|c|c|c|c|c|c|c}
\hline
Method  & B@1 & B@2 & B@3 & B@4 & M & R & C \\
\hline
$\text{GM}_{\textsc{word}}$ & 40.1 & 16.4&6.7 & 2.2 &  15.6 & 28.4 & 9.5 \\
$\text{GM}_{\textsc{word}}$+$\text{CMM}$ &  43.1 &19.4 & 8.3 &  3.0  &  16.5 & 29.4  & 12.6\\
\hline
$\text{GM}_{\textsc{Word+Sub.}}$ &  37.3& 14.9 & 6.1  & 2.5 & 14.3 & 27.0  & 7.9 \\
$\text{GM}_{\textsc{Word+Sub.}}$+$\text{CMM}$& 40.6  &  17.8&7.6& 2.8   & 15.2  &  28.3   & 10.8 \\
\hline
$\text{HGM}_{\textsc{base}}$ & 38.0  & 15.5 & 6.2 &  2.4  & 14.4 & 27.3& 8.0 \\
$\text{HGM}_{\textsc{base}}$+$\text{CMM}$ & 39.8  &  16.9&7.3& 2.6 & 14.8 &  27.7 & 10.2 \\
\hline
$\text{HGM}$  & {41.1} & {17.1} &{6.9}  &  {2.6}  & {15.7} & {28.4} & {9.7} \\
$\text{HGM}$+$\text{CMM}$ & \textbf{44.9}  & \textbf{19.9} &\textbf{8.6} &  \textbf{3.3}  &  \textbf{16.5} & \textbf{29.6} & \textbf{12.7} \\
\hline
\end{tabular}}
\end{center}
\vspace{-2mm}
\caption{ Effectiveness of cross-model feature mapping. Performance on the test split of COCO-CN is used for comparison. C is short for CIDEr.}\label{joint+cyclegan}
\vspace{-2mm}
\end{table}

\subsection{Human Evaluation}
Table~\ref{tab:human_eval} shows the results of human evaluation for baseline models with GAN-based cross-modal feature mapping.
The quality of captions is measured by the \textit{fluency} and \textit{relevancy} metrics. The \textit{fluency} measures whether the generated caption {is} fluent and similar to human-generated captions. The \textit{relevancy} measures whether the caption correctly describes the relevant information of the image. 
These metrics are graded by 5 levels: 1-Very poor, 2-Poor, 3-Adequate, 4-Good, 5-Excellent.
We invite 10 Chinese native speakers to participate in the evaluation, who are from diverse professional backgrounds. Specifically, each participant is randomly assigned with 100 images from COCO-CN test split (1,000 samples in total).
The results in Table~\ref{tab:human_eval} report the mean scores, which illustrate that our method can generate relevant and human-like captions.

\begin{table}[h]
\small
\vspace{-2mm}
\begin{center}
\setlength{\tabcolsep}{1.5pt}{
\begin{tabular}{l|c|c|c|c|c}
\hline
Metric & \shortstack{$\text{GM}_{\textsc{word}}$} & \shortstack{$\text{GM}_{\textsc{word+SUB.}}$} & \shortstack{HGM} &\shortstack{HGM$^\star$} & GT\\
\hline
Relevancy & 2.78 & 2.96 & 3.22 & 3.96 & 4.86\\
Fluency & 2.49 & 2.76 & 3.05 & 4.06 & 4.91\\
\hline
\end{tabular}
}
\end{center}
\vspace{-4mm}
\caption{Human evaluation on COCO-CN test split. HGM$^\star$ represents fine-tuned HGM. Models are trained with CMM. }\label{tab:human_eval}
\vspace{-3mm}
\end{table}

\subsection{ Qualitative Results}\label{sec:qualitative}
\begin{figure}
	\centering
	\includegraphics[width=0.48 \textwidth]{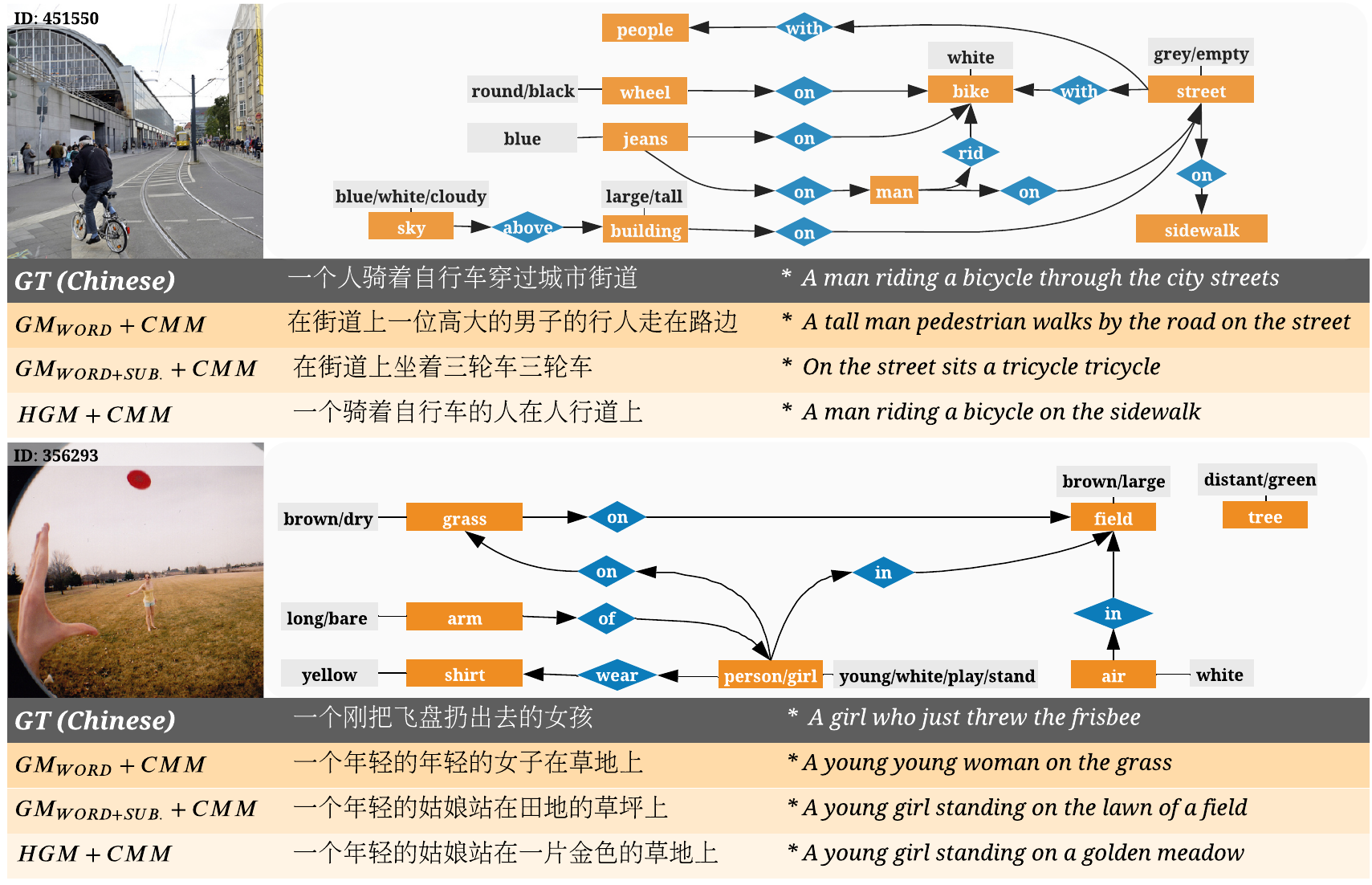}
	\vspace{-4mm}
	\caption{Qualitative results of different unsupervised cross-lingual caption generation models.}
    \vspace{-5mm}
	\label{fig:cocon-example}
\end{figure}

We provide some Chinese captioning examples for MS-COCO images in Fig.~\ref{fig:cocon-example}.
We can see that our method can generate reasonable image descriptions without using any paired image-caption data. Also, we observe that the image scene graphs are quite noisy, which potentially explains the performance degradation when introducing graph-mappings without self-fusion mechanism (see Table~\ref{joint+cyclegan}).
\jh{Figure in Appendix A also visualizes some examples of generated scene graphs with or without augmentation, and its correspondingly generated sentences.}

\section{Conclusion}
In this paper, we {propose} a novel framework to learn a cross-lingual image captioning model without any image-caption pairs.
Extensive experiments demonstrate our proposed methods can achieve promising results for caption generation in the {target} language without using any caption corpus {for training}. We hope our work can provide inspiration for unpaired image captioning in the future.

\section*{Acknowledgments}
We would like to thank Lingpeng Kong, Renjie Pi and the anonymous reviewers for insightful suggestions that have significantly improved the paper. 
This work was supported by TCL Corporate
Research (Hong Kong).
The research of Philip L.H. Yu was supported by a start-up research grant from the Education University of Hong Kong (\#R4162). 

\appendix
\section{Effectiveness of Data Augmentation}
\paragraph{Quantitative Results.}
We investigate the effectiveness of the data augmentation in Table~\ref{tab:aug_performance}. For `$\text{GM}_\textsc{base},\text{Raw}$', we train $\text{GM}_\textsc{base}$ with the original MT corpus. For $\text{GM}_\textsc{base}$, we train the model with the augmented MT corpus.
We can see that model trained on augmented MT corpus achieves better performance.
This is reasonable since the sentence scene graph generated from five sentences (one original sentence and 
four augmented sentences) contains richer information than the original sentence scene graph. 
\begin{figure}[h!]
\centering
	\vspace{-3mm}
	\includegraphics[width=0.48\textwidth]{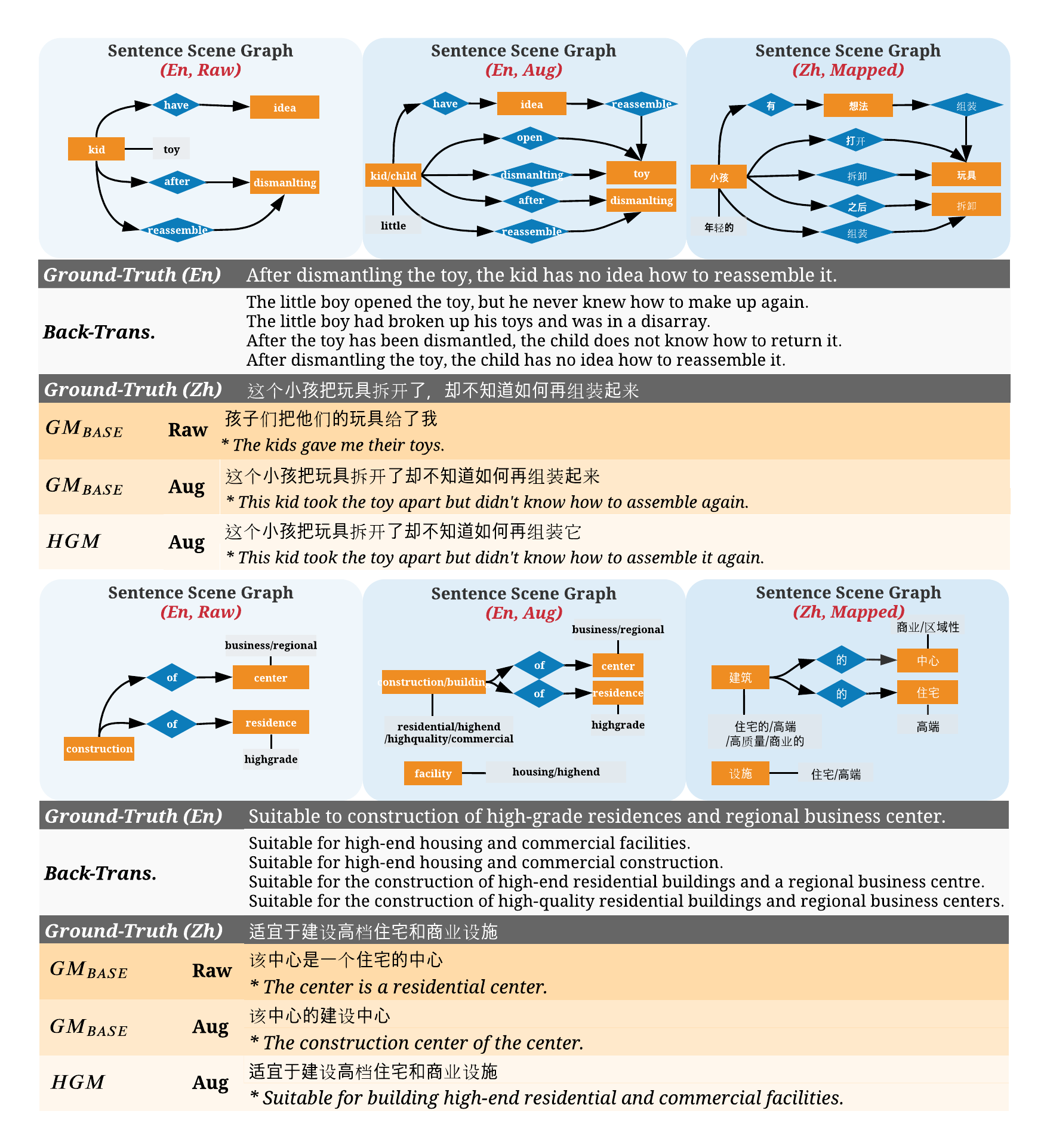}
	\caption{Qualitative results of models trained on different sentence scene graphs. `Raw’ represents that model is trained on the original corpus. `Aug’ represents that model is trained on the augmented corpus. * represents English translation translated by google translator.}
	\label{fig:trans-example}
\end{figure}

\begin{table}[ht]
\small
\vspace{-1.5mm}
\begin{center}
    \setlength{\tabcolsep}{1.5pt}{
\begin{tabular}{l|c|c|c|c|c|c|c}
\hline
Method & $\mathcal{L}_{\text{XE}}$ & $\mathcal{L}_{\text{KL}}$ &$\geqslant$3 Obj/$\mathcal{G}$& B@1 & B@4 & M & R \\
\hline
$\text{GM}_\textsc{base},\text{Raw}$ & \cmark &  \xmark&  15.4\% &22.5 & 5.1 & 12.8  & 24.6 \\
$\text{GM}_\textsc{base}$ & \cmark &  \xmark&  \textbf{59.3\%} & \textbf{26.6} & \textbf{7.3} & \textbf{15.1} & \textbf{28.1} \\
\hline
\end{tabular}
}
\end{center}
\vspace{-2mm}
\caption{Performance comparison of the Chinese sentence generation on the test split of MT corpus. $\geqslant$3 Obj/$\mathcal{G}$ means the percentage of scene graphs whose number of objects is greater than or equal to 3. 
B@$n$ is short for BLEU-$n$, M is short for METEOR and R is short for ROUGE.}\label{tab:aug_performance}
\vspace{-3mm}
\end{table}

\paragraph{Qualitative Results.}
Fig.~\ref{fig:trans-example} also visualizes some augmented scene graphs and examples of generated sentences. Generally, compared with the raw English sentence scene graphs, the English sentence scene graphs generated from the augmented sentences are mucher richer. Also, the HGM model in both figures can generate more accurate and descriptive sentences than other baselines, demonstrating that our hierarchical graph mapping can better map the scene graph from one language to another.

\section{Implementation details}
In this paper, we use English and Chinese as the source and target languages, respectively.We collect a paired English-Chinese corpus from existing MT datasets as the parallel corpus in cross-lingual auto-encoding phase.
For cross-lingual auto-encoding, we collect a paired English-Chinese corpus from existing MT datasets, including WMT19~\cite{barrault2019findings}, AIC\_MT~\cite{wu2017ai}, UM~\cite{tian-etal-2014-um}, and Trans-zh ~\cite{bright_xu_2019_3402023}. 
Since the vocabulary in the collected MT corpus is quite different from the vocabulary in image caption datasets, we filter the sentences in MT datasets according to an {existing caption-style} dictionary {containing} {7,096} words in \citet{li2019coco}.
All the {sentences} longer than 16 or shorter than 10 words are dropped. We further filter the sentences in MT corpus by {reserving} those sentences with 90\% of words in the sentence that appeared in the dictionary, resulting in a filtered MT corpus of 161,613 sentence pairs.

\bibliography{ref}

\begin{thebibliography}{39}
\providecommand{\natexlab}[1]{#1}

\bibitem[{Anderson et~al.(2016)Anderson, Fernando, Johnson, and
  Gould}]{anderson2016spice}
Anderson, P.; Fernando, B.; Johnson, M.; and Gould, S. 2016.
\newblock Spice: Semantic propositional image caption evaluation.
\newblock In \emph{ECCV}.

\bibitem[{Anderson et~al.(2018)Anderson, He, Buehler, Teney, Johnson, Gould,
  and Zhang}]{anderson2018bottom}
Anderson, P.; He, X.; Buehler, C.; Teney, D.; Johnson, M.; Gould, S.; and
  Zhang, L. 2018.
\newblock Bottom-up and top-down attention for image captioning and visual
  question answering.
\newblock In \emph{CVPR}.

\bibitem[{Barrault et~al.(2019)Barrault, Bojar, Costa-Juss{\`a}, Federmann,
  Fishel, Graham, Haddow, Huck, Koehn, Malmasi et~al.}]{barrault2019findings}
Barrault, L.; Bojar, O.; Costa-Juss{\`a}, M.~R.; Federmann, C.; Fishel, M.;
  Graham, Y.; Haddow, B.; Huck, M.; Koehn, P.; Malmasi, S.; et~al. 2019.
\newblock Findings of the 2019 conference on machine translation (wmt19).
\newblock In \emph{WMT}.

\bibitem[{Conneau et~al.(2018)Conneau, Rinott, Lample, Williams, Bowman,
  Schwenk, and Stoyanov}]{Conneau2018xnli}
Conneau, A.; Rinott, R.; Lample, G.; Williams, A.; Bowman, S.; Schwenk, H.; and
  Stoyanov, V. 2018.
\newblock {XNLI}: Evaluating Cross-lingual Sentence Representations.
\newblock In \emph{EMNLP}.

\bibitem[{Denkowski and Lavie(2014)}]{denkowski2014meteor}
Denkowski, M.; and Lavie, A. 2014.
\newblock Meteor universal: Language specific translation evaluation for any
  target language.
\newblock In \emph{ACL}.

\bibitem[{Eberhard, Simons, and Fennig(2019)}]{ethno2019}
Eberhard, D.~M.; Simons, G.~F.; and Fennig, C.~D., eds. 2019.
\newblock \emph{Ethnologue: Languages of the World}.
\newblock SIL International, 22 edition.

\bibitem[{Espl{\`a} et~al.(2019)Espl{\`a}, Forcada, Ram{\'\i}rez-S{\'a}nchez,
  and Hoang}]{espla-etal-2019-paracrawl}
Espl{\`a}, M.; Forcada, M.; Ram{\'\i}rez-S{\'a}nchez, G.; and Hoang, H. 2019.
\newblock {P}ara{C}rawl: Web-scale parallel corpora for the languages of the
  {EU}.
\newblock In \emph{Proceedings of Machine Translation Summit XVII Volume 2:
  Translator, Project and User Tracks}, 118--119. Dublin, Ireland: European
  Association for Machine Translation.

\bibitem[{Feng et~al.(2019)Feng, Ma, Liu, and Luo}]{feng2019unsupervised}
Feng, Y.; Ma, L.; Liu, W.; and Luo, J. 2019.
\newblock Unsupervised image captioning.
\newblock In \emph{CVPR}.

\bibitem[{Gu et~al.(2018)Gu, Joty, Cai, and Wang}]{gu2018unpaired}
Gu, J.; Joty, S.; Cai, J.; and Wang, G. 2018.
\newblock Unpaired image captioning by language pivoting.
\newblock In \emph{ECCV}.

\bibitem[{Gu et~al.(2019)Gu, Joty, Cai, Zhao, Yang, and Wang}]{gu2019unpaired}
Gu, J.; Joty, S.; Cai, J.; Zhao, H.; Yang, X.; and Wang, G. 2019.
\newblock Unpaired image captioning via scene graph alignments.
\newblock In \emph{ICCV}.

\bibitem[{Gu et~al.(2020)Gu, Kuen, Joty, Cai, Morariu, Zhao, and
  Sun}]{gu2020self}
Gu, J.; Kuen, J.; Joty, S.; Cai, J.; Morariu, V.; Zhao, H.; and Sun, T. 2020.
\newblock Self-supervised relationship probing.
\newblock \emph{NeurIPS}.

\bibitem[{Hu et~al.(2020)Hu, Ruder, Siddhant, Neubig, Firat, and
  Johnson}]{hu2020xtreme}
Hu, J.; Ruder, S.; Siddhant, A.; Neubig, G.; Firat, O.; and Johnson, M. 2020.
\newblock XTREME: A Massively Multilingual Multi-task Benchmark for Evaluating
  Cross-lingual Generalization.
\newblock \emph{CoRR}, abs/2003.11080.

\bibitem[{Johnson, Gupta, and Fei-Fei(2018)}]{johnson2018image}
Johnson, J.; Gupta, A.; and Fei-Fei, L. 2018.
\newblock Image generation from scene graphs.
\newblock In \emph{CVPR}.

\bibitem[{Joulin et~al.(2018)Joulin, Bojanowski, Mikolov, J\'egou, and
  Grave}]{joulin2018loss}
Joulin, A.; Bojanowski, P.; Mikolov, T.; J\'egou, H.; and Grave, E. 2018.
\newblock Loss in Translation: Learning Bilingual Word Mapping with a Retrieval
  Criterion.
\newblock In \emph{EMNLP}.

\bibitem[{Krishna et~al.(2017)Krishna, Zhu, Groth, Johnson, Hata, Kravitz,
  Chen, Kalantidis, Li, Shamma et~al.}]{krishna2017visual}
Krishna, R.; Zhu, Y.; Groth, O.; Johnson, J.; Hata, K.; Kravitz, J.; Chen, S.;
  Kalantidis, Y.; Li, L.-J.; Shamma, D.~A.; et~al. 2017.
\newblock Visual genome: Connecting language and vision using crowdsourced
  dense image annotations.
\newblock \emph{IJCV}.

\bibitem[{Laina, Rupprecht, and Navab(2019)}]{laina2019towards}
Laina, I.; Rupprecht, C.; and Navab, N. 2019.
\newblock Towards Unsupervised Image Captioning with Shared Multimodal
  Embeddings.
\newblock In \emph{ICCV}.

\bibitem[{Lan, Li, and Dong(2017)}]{lan2017fluency}
Lan, W.; Li, X.; and Dong, J. 2017.
\newblock Fluency-guided cross-lingual image captioning.
\newblock In \emph{ACMMM}.

\bibitem[{Li et~al.(2019)Li, Xu, Wang, Lan, Jia, Yang, and Xu}]{li2019coco}
Li, X.; Xu, C.; Wang, X.; Lan, W.; Jia, Z.; Yang, G.; and Xu, J. 2019.
\newblock COCO-CN for Cross-Lingual Image Tagging, Captioning, and Retrieval.
\newblock \emph{TMM}.

\bibitem[{Lin(2004)}]{lin2004rouge}
Lin, C.-Y. 2004.
\newblock Rouge: A package for automatic evaluation of summaries.
\newblock In \emph{ACL}.

\bibitem[{Lin et~al.(2014)Lin, Maire, Belongie, Hays, Perona, Ramanan,
  Doll{\'a}r, and Zitnick}]{lin2014microsoft}
Lin, T.-Y.; Maire, M.; Belongie, S.; Hays, J.; Perona, P.; Ramanan, D.;
  Doll{\'a}r, P.; and Zitnick, C.~L. 2014.
\newblock Microsoft coco: Common objects in context.
\newblock In \emph{ECCV}.

\bibitem[{Ng et~al.(2019)Ng, Yee, Baevski, Ott, Auli, and
  Edunov}]{ng2019facebook}
Ng, N.; Yee, K.; Baevski, A.; Ott, M.; Auli, M.; and Edunov, S. 2019.
\newblock Facebook FAIR's WMT19 News Translation Task Submission.
\newblock \emph{arXiv preprint arXiv:1907.06616}.

\bibitem[{Nguyen et~al.(2021)Nguyen, Tripathi, Du, Guha, and
  Nguyen}]{Nguyen_2021_ICCV}
Nguyen, K.; Tripathi, S.; Du, B.; Guha, T.; and Nguyen, T.~Q. 2021.
\newblock In Defense of Scene Graphs for Image Captioning.
\newblock In \emph{ICCV}.

\bibitem[{Papineni et~al.(2002)Papineni, Roukos, Ward, and
  Zhu}]{papineni2002bleu}
Papineni, K.; Roukos, S.; Ward, T.; and Zhu, W.-J. 2002.
\newblock BLEU: a method for automatic evaluation of machine translation.
\newblock In \emph{ACL}.

\bibitem[{Ren et~al.(2015)Ren, He, Girshick, and Sun}]{ren2015faster}
Ren, S.; He, K.; Girshick, R.; and Sun, J. 2015.
\newblock Faster r-cnn: Towards real-time object detection with region proposal
  networks.
\newblock In \emph{NeurIPS}.

\bibitem[{Rennie et~al.(2017)Rennie, Marcheret, Mroueh, Ross, and
  Goel}]{rennie2017self}
Rennie, S.~J.; Marcheret, E.; Mroueh, Y.; Ross, J.; and Goel, V. 2017.
\newblock Self-critical sequence training for image captioning.
\newblock In \emph{CVPR}.

\bibitem[{Schuster et~al.(2015)Schuster, Krishna, Chang, Fei-Fei, and
  Manning}]{schuster2015generating}
Schuster, S.; Krishna, R.; Chang, A.; Fei-Fei, L.; and Manning, C.~D. 2015.
\newblock Generating semantically precise scene graphs from textual
  descriptions for improved image retrieval.
\newblock In \emph{ACL}.

\bibitem[{Song et~al.(2019)Song, Chen, Zhao, and Jin}]{song2019unpaired}
Song, Y.; Chen, S.; Zhao, Y.; and Jin, Q. 2019.
\newblock Unpaired Cross-lingual Image Caption Generation with Self-Supervised
  Rewards.
\newblock In \emph{ACMMM}.

\bibitem[{Tian et~al.(2014)Tian, Wong, Chao, Quaresma, Oliveira, and
  Yi}]{tian-etal-2014-um}
Tian, L.; Wong, D.~F.; Chao, L.~S.; Quaresma, P.; Oliveira, F.; and Yi, L.
  2014.
\newblock UM-Corpus: A Large English-Chinese Parallel Corpus for Statistical
  Machine Translation.
\newblock In \emph{LREC}.

\bibitem[{Tran et~al.(2016)Tran, He, Zhang, Sun, Carapcea, Thrasher, Buehler,
  and Sienkiewicz}]{tran2016rich}
Tran, K.; He, X.; Zhang, L.; Sun, J.; Carapcea, C.; Thrasher, C.; Buehler, C.;
  and Sienkiewicz, C. 2016.
\newblock Rich image captioning in the wild.
\newblock In \emph{CVPRW}.

\bibitem[{Vedantam, Lawrence~Zitnick, and Parikh(2015)}]{vedantam2015cider}
Vedantam, R.; Lawrence~Zitnick, C.; and Parikh, D. 2015.
\newblock Cider: Consensus-based image description evaluation.
\newblock In \emph{CVPR}.

\bibitem[{Vinyals et~al.(2015)Vinyals, Toshev, Bengio, and
  Erhan}]{vinyals2015show}
Vinyals, O.; Toshev, A.; Bengio, S.; and Erhan, D. 2015.
\newblock Show and tell: A neural image caption generator.
\newblock In \emph{CVPR}.

\bibitem[{Wang et~al.(2018)Wang, Liu, Zeng, and Yuille}]{wang2018scene}
Wang, Y.-S.; Liu, C.; Zeng, X.; and Yuille, A. 2018.
\newblock Scene graph parsing as dependency parsing.
\newblock \emph{arXiv preprint arXiv:1803.09189}.

\bibitem[{Wu et~al.(2017)Wu, Zheng, Zhao, Li, Yan, Liang, Wang, Zhou, Lin, Fu
  et~al.}]{wu2017ai}
Wu, J.; Zheng, H.; Zhao, B.; Li, Y.; Yan, B.; Liang, R.; Wang, W.; Zhou, S.;
  Lin, G.; Fu, Y.; et~al. 2017.
\newblock AI Challenger: A Large-scale Dataset for Going Deeper in Image
  Understanding.
\newblock \emph{arXiv preprint arXiv:1711.06475}.

\bibitem[{Wu et~al.(2016)Wu, Schuster, Chen, Le, Norouzi, Macherey, Krikun,
  Cao, Gao, Macherey et~al.}]{wu2016google}
Wu, Y.; Schuster, M.; Chen, Z.; Le, Q.~V.; Norouzi, M.; Macherey, W.; Krikun,
  M.; Cao, Y.; Gao, Q.; Macherey, K.; et~al. 2016.
\newblock Google's neural machine translation system: Bridging the gap between
  human and machine translation.
\newblock \emph{arXiv preprint arXiv:1609.08144}.

\bibitem[{Xu(2019)}]{bright_xu_2019_3402023}
Xu, B. 2019.
\newblock NLP Chinese Corpus: Large Scale Chinese Corpus for NLP.

\bibitem[{Yang et~al.(2019)Yang, Tang, Zhang, and Cai}]{yang2019auto}
Yang, X.; Tang, K.; Zhang, H.; and Cai, J. 2019.
\newblock Auto-encoding scene graphs for image captioning.
\newblock In \emph{CVPR}.

\bibitem[{Zellers et~al.(2018)Zellers, Yatskar, Thomson, and
  Choi}]{zellers2018neural}
Zellers, R.; Yatskar, M.; Thomson, S.; and Choi, Y. 2018.
\newblock Neural motifs: Scene graph parsing with global context.
\newblock In \emph{CVPR}.

\bibitem[{Zhong et~al.(2020)Zhong, Wang, Chen, Yu, and
  Li}]{DBLP:conf/eccv/ZhongWC0L20}
Zhong, Y.; Wang, L.; Chen, J.; Yu, D.; and Li, Y. 2020.
\newblock Comprehensive Image Captioning via Scene Graph Decomposition.
\newblock In Vedaldi, A.; Bischof, H.; Brox, T.; and Frahm, J., eds.,
  \emph{ECCV}.

\bibitem[{Zhu et~al.(2017)Zhu, Park, Isola, and Efros}]{zhu2017unpaired}
Zhu, J.-Y.; Park, T.; Isola, P.; and Efros, A.~A. 2017.
\newblock Unpaired image-to-image translation using cycle-consistent
  adversarial networks.
\newblock In \emph{ICCV}.

\end{thebibliography}

\end{document}